\documentclass[11pt,a4paper]{article}
\usepackage[hyperref]{acl2020}
\usepackage{times}
\usepackage{latexsym}

\usepackage{textcomp}
\usepackage{microtype}
\usepackage{subcaption}
\usepackage{graphicx}
\usepackage{multirow}
\usepackage[utf8]{inputenc}
\usepackage{caption}
\aclfinalcopy 

\title{Character-level Transformer-based Neural Machine Translation}

\author{Nikolay Banar\textsuperscript{1,2,3} \And
  Walter Daelemans\textsuperscript{1,2}  \\
  \textsuperscript{1}University of Antwerp, Belgium  \\
  \textsuperscript{2}Computational Linguistics and Psycholinguistics Research Center, Belgium \\
 \textsuperscript{3}Antwerp Centre for Digital Humanities and Literary Criticism, Belgium \\
  \normalsize  \texttt{\texttt{\{}nicolae.banari, walter.daelemans, mike.kestemont}\texttt{\}email@uantwerpen.be} \\ \\
  \\\And
  Mike Kestemont\textsuperscript{1,2,3} \\
}

\date{}

\begin{document}
\maketitle
\begin{abstract}
Neural machine translation (NMT) is nowadays commonly applied at the subword level, using byte-pair encoding. A promising alternative approach focuses on character-level translation, which simplifies processing pipelines in NMT considerably. This approach, however, must consider relatively longer sequences, rendering the training process prohibitively expensive. In this paper, we discuss a novel, Transformer-based approach, that we compare, both in speed and in quality to the Transformer at subword and character levels, as well as previously developed character-level models. We evaluate our models on 4 language pairs from WMT’15: DE-EN, CS-EN, FI-EN and RU-EN. The proposed novel architecture can be trained on a single GPU and is $\sim$34\% faster than the character-level Transformer; still, the obtained results are at least on par with it. In addition, our proposed model outperforms the subword-level model in FI-EN and shows close results in CS-EN. To stimulate further research in this area and close the gap with subword-level NMT, we make all our code and models publicly available.
\end{abstract}

\section{Introduction}

Sequence-to-sequence models are nowadays a mainstream approach in Neural Machine Translation (NMT). Such models are typically applied at the subword level based on byte-pair encoding (BPE), originally proposed by~\citet{sennrich2015neural}. This algorithm mitigates the problem of rare and out-of-vocabulary words that present a significant issue for word-level models. BPE builds a vocabulary of the most frequent subword units of different lengths, starting from a single character. Then, the input sentence is divided into a sequence of the longest possible subword fragments matching the constructed vocabulary. This approach is appealing because of its strong empirical results and computational efficiency. However, the segmentation is language- and corpus-dependent and, hence, requires considerable hyperparameter tuning. The problem of finding an optimal subword segmentation is especially challenging for multilingual and zero-short translation~\cite{johnson2017google}.

Another recent direction in NMT focuses on character-level translation. This approach is conceptually attractive because it can help mitigate the previously mentioned shortcomings of subword-level models. Character-level models do not rely on an explicit segmentation of the input sentence (be it rule-based or statistical) and resort to plain characters as a sentence's basic units. As such, models are implicitly enforced to learn the inner structure of complex words. Hence, such models are more robust in the face of out-of-vocabulary words and in translating noisy and out-of-domain text. In comparison to subword-level models, they should be able to model more accurately rare morphological variants of words \cite{chung2016character,lee2017fully, gupta2019character}. In addition, character-level models may work better in some fine-tuning scenarios, where the amount of available data is challengingly small \cite{banar2020}.

In spite of its conceptual elegance, the character-level approach also presents considerable challenges, that help explain why this approach didn't receive much attention yet. Character sequences are significantly longer and, consequently, more challenging to model. Moreover, the level of semantics in character-level representation becomes even more abstract and, hence, larger models with a highly non-linear mapping function are required. Finally, the training and decoding time for such models is much longer. However, some of these issues can be tackled through resorting to new NMT architectures. \citet{lee2017fully} have shown that is possible to train a character-level model, within a reasonable time span, by reducing the length of the source representation. We utilize this publicly available model, henceforth: CharRNN, as a baseline in our experiments.

We base our work on the well-known Transformer architecture \citet{vaswani2017attention}, which has shown state-of-the-art performance on several language pairs in NMT. The model is intrinsically very attractive for the character level due to the high training speed it enables and its strong modelling capacity with respect to longer-range dependencies. The Transformer relies on self-attention and does not include any recurrence in training. Therefore, the Transformer can be fully parallelized during training, leading to considerable speed-ups in comparison to recurrent networks.

We aim to stimulate further research in this direction, by demonstrating the computational feasibility of training fast character-level models, even on a single GPU. Below, we propose a new variant (CharTransformer) of a publicly available, Transformer-based network and apply it at the character level. Our models applies the same source length reduction technique as \citet{lee2017fully} and introduces a six-layer Transformer at the encoder and decoder sides instead of recurrent layers as in CharRNN, making our network fully parallelizable. The main contribution of the paper is two-fold: (i) We demonstrate the feasibility of training high-quality and fast character-level translation models, even on a single GPU; (ii) we propose a novel character-level Transformer-based architecture that is at least as accurate as the Transformer, yet is up to ~34\% faster.

\section{Related Work}
In this section, we survey recent work in the field of character-level NMT that is directly relevant to the present paper. \citet{costa2016character} utilized a convolutional network to extract local dependencies from character embeddings and, downstream, applied a Highway network~\cite{srivastava2015training} to construct segmented embeddings. This model showed promising results but, crucially, still relied on a word-level segmentation at the decoder and encoder sides. \citet{ling2015character} assembled word embeddings from character embeddings via bidirectional long short-term memory units (LSTM, ~\citet{hochreiter1997long}). The model decoded the target words character-by-character and outperformed a comparable word-based baseline. However, the training time was substantially longer and, still, explicit segmentation was required.

\citet{luong2016achieving} used character-level information to mitigate out-of-vocabulary issues in a word-based model. Additionally, they compared a fully character-level model with a word-level baseline. Notwithstanding comparable results, the fully character-level model was significantly slower. \citet{chung2016character} compared character-level and subword-level decoders, while the encoder still worked at the subword level. Their experiments demonstrated that the character-level decoder could outperform the subword-level one.

\citet{lee2017fully} were the first to propose a fully character-level model that came with computational requirements comparable to those of subword-level models. At the encoder side, they efficiently reduced the length of the input sequences via the use of a convolutional layer, a max-pooling layer and a stack of Highway layers. On top of the encoder, they used bidirectional gated recurrent units (GRU,~\citet{cho2014learning}). In this paper too, the character-level NMT model was able to outperform the subword-level baseline. Finally, and in the same spirit, \citet{cherry2018revisiting} showed that standard character-level models of sufficient depth are able to outperforms comparable subword-level models. However, they utilized a prohibitively expensive training regime with 16 GPUs (training times were not explicitly reported for each network) and did not make their models publicly available. Hence, we do not consider these models below and restrict ourselves to publicly available implementations. \citet{gupta2019character} demonstrated that the character-level Transformer is competitive to the subword-level Transformer, but does not outperform it.

Here, we take inspiration from \citet{chen2018best}, who investigated different NMT architectures, including hybrid models with Transformers. They demonstrated the superiority of the Transformer encoder over the recurrent encoder at the subword level. We hypothesize that the CharRNN model may be easily improved by incorporating the Transformer approach, instead of the more conventional, recurrent layers. In addition, the architecture can be sped up at the training phase by using the Transformer decoder (as in CharTransformer). Our work is therefore the first to assess the effectiveness and efficiency of CharTransformer.  

\section{Background}

\begin{table}[t!]
\begin{center}
\begin{tabular}{|l|r|r|}
\hline \multicolumn{3}{|c|}{Encoder}  \\\hline
\textbf{Param.} & \textbf{Transformer} & \textbf{CharTrans.} \\ \hline
Emb. & 512 & 128 \\\hline
Conv. &  & 200-200-250-250\\
filters &  & 300-300-300-300 \\\hline
Pool stride &  & 5 \\\hline
Highway &  & 2 \\\hline
Layers &  \multicolumn{2}{c|}{6}  \\\hline 
$d_{m}, d_{k}, d_{v}$ &  \multicolumn{2}{c|}{512}  \\\hline 
Heads &  \multicolumn{2}{c|}{8}  \\\hline 
$d_{ff}$ &  \multicolumn{2}{c|}{2048}  \\\hline 
\multicolumn{3}{c}{}  \\\hline
\multicolumn{3}{|c|}{Decoder}  \\\hline
\textbf{Param.} & \textbf{Transformer} & \textbf{CharTrans.} \\ \hline
Emb. &  \multicolumn{2}{c|}{512} \\\hline
Layers &  \multicolumn{2}{c|}{6}  \\\hline 
$d_{m}, d_{k}, d_{v}$ &  \multicolumn{2}{c|}{512}  \\\hline 
Heads &  \multicolumn{2}{c|}{8}  \\\hline 
$d_{ff}$ &  \multicolumn{2}{c|}{2048}  \\\hline 

\end{tabular}
\end{center}
\caption{\label{encoder} Encoder and decoder parameters of the investigated models. At the encoder side, the models utilize 200 filters of width 1, 200 filters of width 2 etc. $d_{ff}$ corresponds to the inner-layer has dimensionality. $d_{m}$ corresponds to the dimensionality of input and output. $d_{k}, d_{v}$ correspond to the dimensionality of keys and values for attention heads, respectively.}
\end{table}

In this section, we briefly discuss two of the commonly used architectures in NMT.

\subsection{Recurrent Neural Networks}
Recurrent models nowadays generally utilize GRU or LSTM memory cells, and follow the encoder-decoder paradigm. They consist of an encoder and an (attentional) decoder \cite{bahdanau2014neural, sutskever2014sequence, luong2015effective, cho2014properties}. The encoder processes a source sentence and constructs a continuous representation of it, which is sometimes considered a summarized meaning of the input sentence. The decoder generates the output sentence. These models are usually trained by minimizing the negative conditional log-likelihood of outputs given the corresponding source sentences and the previously observed target tokens. \\

\noindent \textbf{Encoder} The encoder processes a source sentence step by step and the current state of the encoder depends on its previous hidden state. A common practice is to apply bidirectional recurrent layers. A forward recurrent layer processes the input sequence from left to right and a backward recurrent layer processes it from right to left. Further, the outputs of the layers are concatenated in order to assemble the final source sentence representation.\\

\noindent \textbf{Attentional Decoder} Depending on the specific architecture, the input of the decoder may include the previously generated token, its previous hidden states and the the context vector. The context vector is built by the attention mechanism. It searches parts of the source sentence that are relevant for each decoding time step.  The context vector is calculated as a weighted sum of the source hidden states. Hence, the weights represent an importance of the input tokens given the current target token. \\

\subsection{Transformer}
\begin{figure*}[h!]
  \centering
    \includegraphics[width=0.4\textwidth]{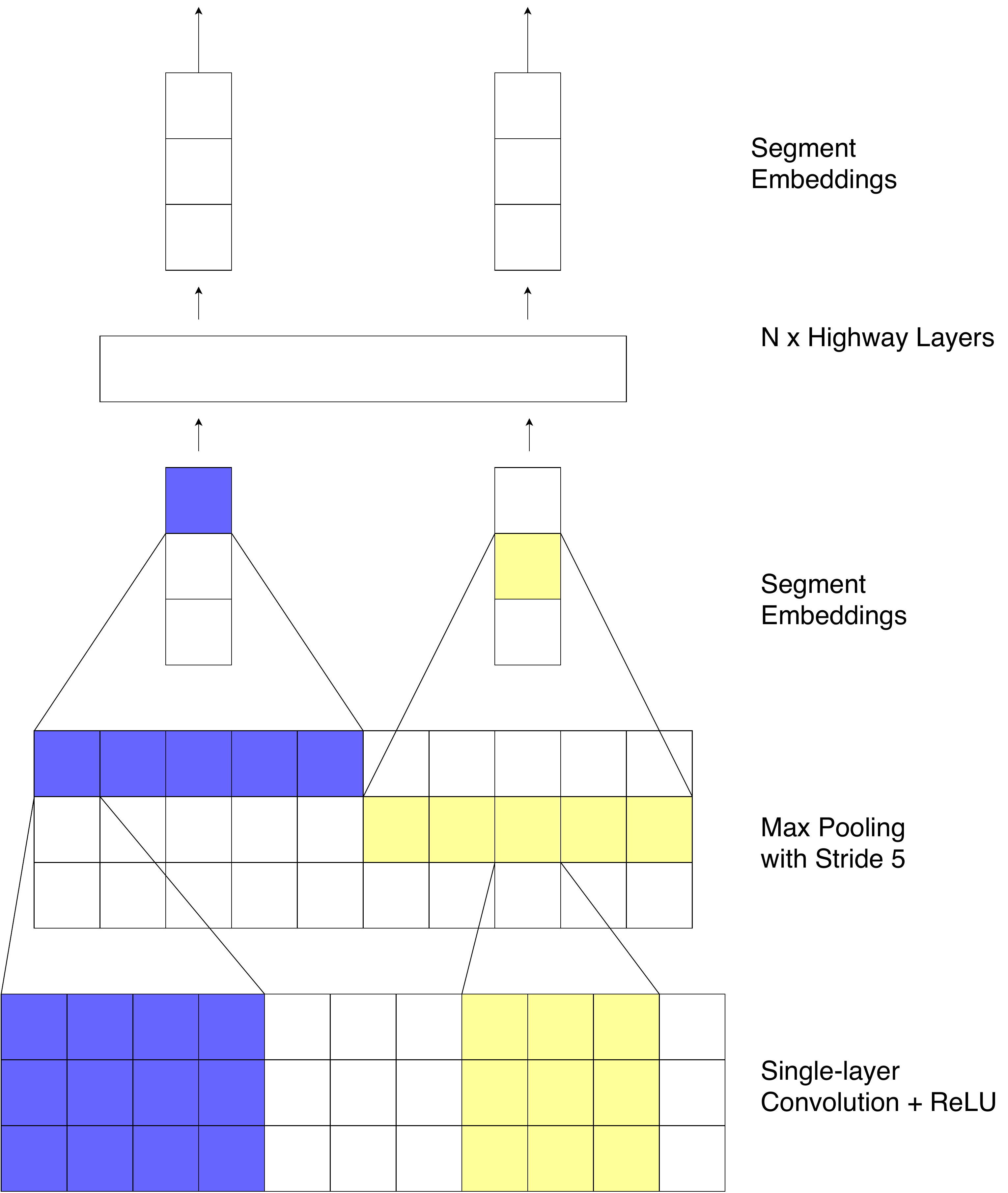}
    \caption{Scheme of the source length reduction technique.}
    \label{EncoderScheme}
\end{figure*}
The Transformer \cite{vaswani2017attention} model aims to overcome some of the issues induced by recurrent and convolutional sequence-to-sequence models. Compared to convolutional models, which have a limited receptive field, the Transformer utilizes self-attention networks. Thereby, the model is able to access all position of the previous layer. In addition, the Transformer does not have any recurrent connections at the training phase that allows to make training process fully parallel. These NMT models still rely on encoder-decoder scheme, which follows the same purpose as for recurrent networks.  Transformers are commonly trained using the Noam decay schedule \cite{popel2018training}, also by minimizing the negative conditional log-likelihood. \\

\noindent \textbf{Encoder} The encoder processes the full sequence simultaneously, as opposed to recurrent approaches. It starts with a positional encoding and processes the full sequence at once. As the Transformer contains no recurrence and no convolution, this step is required to provide information about the position of the tokens in the sequence. The encoder in each layer consists of 2 sub-layers: a self-attention network and a feed-forward neural network. In addition, a residual connection around each sub-layer is utilized. Downstream, layer normalization is applied. The encoder, because of its immediacy, is fully parallelizable in training and decoding phases.\\

\noindent  \textbf{Decoder} In comparison to the encoder, decoder layers have an additional self-attention network between 2 sub-layers that attend to the encoder. The decoder is fully parallelizable in the training phase. However, decoding is conducted step by step similarly to recurrent networks.\\

\section{Machine Translation Models}

In this work, we compare three character-level and one subword-level NMT systems. First, we report results for the character-level model proposed by \citet{lee2017fully} and use it as a baseline (CharRNN). In this model, the decoder consists of two unidirectional GRU layers and the attention score is computed by a single-layer feedforward network. The encoder part implements an efficient source length reduction technique (detailed below), and adds a single-layer, bidirectional GRU on top. Second, we train a character-level Transformer and a subword-level Transformer \cite{vaswani2017attention} without any architectural modifications. And finally, we apply the source length reduction technique to the Transformer and build CharTransformer. We implemented this model in PyTorch \cite{NEURIPS2019_9015}, inside the OpenNMT-py framework \cite{opennmt}. Further information about the parameters of the encoders and the decoders of the Transformer and CharTransformer are summarized in Table~\ref{encoder}. Layer sizes of the models are kept maximally comparable. Below, we highlight the important details of the models.

\subsection{Source Length Reduction}
As a recurrent baseline model, we use the model proposed by \citet{lee2017fully}. The encoder employs one-dimensional convolutions, following with max-pooling layers and a Highway network, in order to reduce the substantial length (up to 450 characters) of the input sentence by a factor of 5 and efficiently construct representation of local features. We briefly highlight the main properties of the source length reduction technique below, which is schematically depicted in Figure~\ref{EncoderScheme}. \\

\noindent  \textbf{Embedding layer} The embedding layer takes the form of a lookup table, which maps a sequence of source tokens to a sequence of embeddings in order to build a continuous representation of each token.\\

\noindent  \textbf{Convolutions} One-dimensional convolutional filters (with padding) are applied to the sequence of the input embeddings produced by the embedding layer. Filter widths range from 1 to 8, which allows to construct representation of n-grams up to 8 characters. Downstream, the outputs of the convolutional filters are stacked and the rectified linear activation is applied.\\

\noindent  \textbf{Max pooling} Conventional max pooling is applied to non-overlapping parts of the convolutional layer output. Thus, the layer reduces the length of the source representation and constructs segment embeddings, containing the most salient features of the source sub-sequences.\\

\noindent  \textbf{Highway layers} The Highway network is introduced after the convolutional part of the encoder. Highway layers~\cite{srivastava2015training} have been shown to improve the quality of character-level models~\cite{kim2016character}.\\

\subsection{CharTransformer Encoder}
In the CharTransformer encoder, we implement the source length reduction technique from \citet{lee2017fully} (Figure~\ref{EncoderScheme}) and inherit the following layers from the baseline: the embedding layer, the convolution layer, the max pooling, the Highway network. On the top of the encoder, we employ a six-layer Transformer. \\

\section{Experimental Settings}
 Below, we provide details of our experiments.

\subsection{Datasets and Preprocessing}
We applied the NMT models to the four language pairs from WMT’15: DE-EN, CS-EN, FI-EN and RU-EN. We obtained the datasets\footnote{\url{https://github.com/nyu-dl /dl4mt-c2c}} already preprocessed by \citet{lee2017fully}, using a script from Moses\footnote{\url{https://github.com/moses-smt/mosesdecoder}}. Although this step is not strictly required for character-level translation, we kept it for the sake of comparison. In addition, we created a tokenized dataset, using another reference routine, \citet{sennrich2015neural}, with 20,000 BPE operations for each of the source and target corpora. We allowed a vocabulary size of 300 tokens for the character-level translation and 20k$--$24k tokens for the subword-level models. We limit the length of sentences to 450 characters or 50 subword tokens. For the FI-EN language pair, we utilized newsdev-2015 as a development set and newstest-2015 as a test set. For other language pairs, we used newstest-2013 as a development set and the combination of newstest-2014 and newstest-2015 as test sets.\\

\subsection{Metrics}
Notwithstanding its reliability, human assessment in machine translation is expensive and slow to obtain. In NMT, a number of automated metrics have therefore been proposed to measure the performance of models. Generally speaking, these measure the quality of a system's output by comparing it to human judgments. Recently, character-level metrics demonstrated the best performance among the non-trainable metrics in the field \cite{ma2018results}. Therefore, we utilized not only the popular metric \textsc{BLEU-4}~\cite{papineni2002bleu}, but also \textsc{CharacTER}\footnote{\url{https://github.com/rwth-i6/CharacTER}}~\cite{wang2016character} and \textsc{CHRF}\footnote{\url{https://github.com/m-popovic/chrF}}~\cite{popovic2015chrf}.

\begin{table*}[ht]
\centering
\begin{tabular}{|l|l|l|r|r|r|r|r|r|}
  \hline
   \multirow{2}{*}{ Lang.} & \multirow{2}{*}{Model} & \multirow{2}{*}{Seg.} & \multicolumn{3}{c|}{Test1}  & \multicolumn{3}{c|}{Test2} \\ \cline{4-9}
   &  &  & BLEU$\uparrow$ & C-TER$\downarrow$ &   CHRF$\uparrow$  & BLEU$\uparrow$ & C-TER$\downarrow$ & CHRF$\uparrow$\\
  \hline
   \multirow{4}{*}{DE-EN} & CharRNN & char & 25.77 & NA & NA &  25.83 & NA & NA\\
  & Transformer & char & 28.32 & 47.41 &  53.14 &  28.70 & 45.44 & 53.08\\
   & CharTransformer & char & 28.63 & 46.54 & 53.70 & 28.08 & \textbf{45.16} & 53.18 \\
  & Transformer & bpe &  \textbf{29.72} &  \textbf{46.35} &  \textbf{54.26} &   \textbf{29.76} & 45.36 &  \textbf{54.11}\\ \hline
  \multirow{4}{*}{CS-EN} & CharRNN & char & 24.08 & NA & NA &  22.46 & NA & NA\\
    & Transformer & char & 24.77 & 48.13 &  50.91 & 23.51 & 51.34 & 48.20\\
      & CharTransformer & char & 26.89 & \textbf{45.40} & 53.66 & 25.24 & \textbf{49.44} & 50.47 \\
  & Transformer & bpe &  \textbf{28.41} & 45.62 &   \textbf{54.02} &   \textbf{26.14} & 49.92 &  \textbf{50.56} \\ \hline
   \multirow{4}{*}{FI-EN} & CharRNN & char & NA & NA & NA &  13.10 & NA  & NA \\
   & Transformer & char & NA & NA &  NA & \textbf{18.72} & \textbf{55.95} & \textbf{44.97}\\
    & CharTransformer & char & NA & NA &  NA &  17.52 & 57.70 & 43.46 \\ 
  & Transformer & bpe & NA & NA & NA  &  17.35 & 58.21 & 42.90\\ \hline
    \multirow{4}{*}{RU-EN} & CharRNN & char & 26.80 & NA & NA &  22.73 & NA & NA\\
    & Transformer & char & 30.87 & \textbf{42.55} &  \textbf{56.80} & 26.99 & \textbf{46.17} & 52.96 \\
     & CharTransformer & char & 30.31 & 42.78 & 56.35 & 26.19 & 46.72 & 52.21 \\ 
  & Transformer & bpe & \textbf{31.39} & 43.21 & 56.75 &  \textbf{28.01} & 46.40 & \textbf{53.41}\\ \hline
\end{tabular}

\caption{Results of the models on 4 language pairs. The best performing models are shown in bold. Results for CharRNN are obtained from \citet{lee2017fully}. The subword-level models are not used in comparison. } 
  \label{table:Res}
\end{table*}

\subsection{Training Details}
We mostly followed the settings, recommended by the OpenNMT-py framework\footnote{\url{https://opennmt.net/OpenNMT-py/FAQ.html}}. The models were trained by minimizing the negative conditional log-likelihood using the Adam optimizer~\cite{kingma2014adam} with an initial learning rate of 2 and the Noam decay schedule \cite{popel2018training}. The models were initialized using the method proposed by \citet{glorot2010understanding}. We did not change any settings for the subword-level models. Below, the parameters that we altered for the character-level models are explicitly listed. As character tokens contain less information compared to subwords, we utilized a larger batch size of 6144 tokens and an accumulation count of 4, to get a more faithful gradient approximation. Additionally, we set dropout to 0 to make the models converge faster. We used -max\_generator\_batches with default parameters. We trained the models for 100,000 updates. Each model was trained on a single GeForce GTX 1080 Ti with 11 GB of memory.\\

\subsection{Encoding Details} 
We slightly altered the implementation of the original source length reduction used by \citet{lee2017fully} in CharRNN to reduce the memory consumption of the model. Highway layers significantly improve the performance of character-level language models based on convolutional networks. Even though, the Highway layers significantly improve the performance of convolution based character-level language models, \citet{kim2016character} demonstrated that they saturate in performance after 2 layers. Therefore, we utilized only 2 (instead of the original 4) layers in CharTransformer to reduce the complexity of the models under consideration.\\

\subsection{Decoding Details} 
In the decoding part, we utilized beam search with beam size of 20 for character-level models and beam size of 5 for subword-level models.

\section{Results and Discussion}
\subsection{Quantitative Analysis}
\begin{table}[t!]
\begin{center}
\begin{tabular}{|l|r|r|r|}
\hline \textbf{Model} & \textbf{Speed} &  \textbf{Overall} & \textbf{Percent}\\ \hline
Trans. & 1,362 & 37,71 & 100 \\ \hline
CharTrans. & 0,894 & 24,76 & 66 \\ \hline

\end{tabular}
\end{center}

\caption{Speed comparison for the character-level models. The second column shows the time of one update in seconds. The third column reports the total training time in hours. The last column shows speed difference in percents. The models make one update after processing four batches.} \label{table:time}
\end{table}

\begin{table}[t!]
\begin{center}
\begin{tabular}{|l|r|r|r|r|}
\hline Metric & \textbf{FI-} & \textbf{DE-} &  \textbf{CS-} & \textbf{RU-}\\ \hline
C-TER & 0.888 &  \textbf{0.972} & 0.960 & 0.884 \\ \hline
CHRF  & 0.903 & 0.956 &  \textbf{0.968} &  \textbf{0.898} \\ \hline
BLEU &  \textbf{0.929} & 0.865 & 0.957 & 0.851 \\ \hline

\end{tabular}
\end{center}

\caption{ WMT15 system-level correlations of automatic evaluation metrics and the official human scores for -EN \cite{wang2016character}. The best results are in bold. } \label{table:metrics}
\end{table}

\begin{table*}[!ht]
\caption* { (a) Named Entities and transliteration ({Russian$\rightarrow$English })}
\begin{tabular*}{\textwidth}{l|l}
\hline
  transliteration & Ostaviv ej golosovoe soobshhenie 18 ijunja 2005-go , \textbf{Koulson} skazal :  [...]  \\ \hline
  target & Leaving the voice message on June 18 , 2005 , \textbf{Caulsen} said : ' [...] \\\hline
  CharRNN & Having left her voicemail on 18 June 2005 , \textbf{Coleson} said , ' [...] \\ \hline
  Transformer (char) & Having left her voicemail on 18 June 2005 , \textbf{Coleson} said , ' [...] \\ \hline
  CharTransformer & Leaving her voice message on June 18 , 2005 , \textbf{Cowlson} said , ' [...] \\\hline
   Transformer (bpe) & Leaving her a voice message on 18 June 2005 , \textbf{Colson} said , ' [...] \\\hline
\end{tabular*}
\bigskip
\caption* {(b) Flooding of chunks and incomplete words  ({Russian$\rightarrow$English })}
\begin{tabular*}{\textwidth}{l|l}
\hline
   transliteration &  Sirija unichtozhila oborudovanie dlja himoruzhija  \\ \hline
  target &  Syria destroyed equipment for chemical weapons \\ \hline
  CharRNN &  Syria destroyed the \textbf{equipment} for the \textbf{equipment} for \textbf{chemothera} \\\hline
  Transformer (char) &  Syria has destroyed chemo-weapons equipment \\ \hline
  CharTransformer & Syria Destroyed Chemical Equipment \\\hline
  Transformer (bpe) &  Syria Destructed Chemical Weapons  \\\hline
\end{tabular*}
\bigskip
\caption* {(c) Fixed expressions ({Russian$\rightarrow$English })}
\begin{tabular*}{\textwidth}{l|l}
\hline
  transliteration &  V Kineshme i rajone dvoe muzhchin pokonchili zhizn' samoubijstvom  \\ \hline
  target & In Kineshma and environs two men have committed suicide \\\hline
  CharRNN & In Kineshma and the area \textbf{of} two men \textbf{committed suicide behavior} \\ \hline
  Transformer (char) & In Kineshma and the region , two men have committed suicide . \\ \hline
  CharTransformer & In Kineshma and the region , two men have ended their lives \textbf{of suicide} \\ \hline
  Transformer (bpe) & In Kineshma and environs two men have committed suicide \\ \hline
\end{tabular*}
\bigskip
\caption* {(d) Conciseness of Transformer ({Russian$\rightarrow$English })}
\begin{tabular*}{\textwidth}{l|p{12.4cm}}
\hline
   transliteration &  Ko vremeni podvedenija itogov tendera byla opredelena arhitekturnaja koncepcija ajerovokzal'nogo kompleksa ' Juzhnyj ' , kotoruju  razrabotala   britanskaja kompanija Twelve Architects
    \\\hline
  target & By the time the tender results were tallied , the architectural concept of the ' Yuzhniy ' air terminal complex , which was developed by the British company Twelve Architects , had been determined . \\\hline
  CharRNN & By the time the tender 's results were defined an architectural concept of the ' South ' architecture complex , which was developed by the British company Twelve Architects . \\ \hline
  Transformer (char) & By the time of summing up the results of the tender the architectural concept of the Yuzhny terminal complex was developed by Twelve Architects . \\ \hline
  CharTransformer & By the time of the summing up of the tender , the architectural concept of the \\ 
   & ' South ' terminal complex developed by the British company Twelve Architects\\ 
   & was identified . \\ \hline
   Transformer (bpe) & By the time the tender results were summed up the architectural concept of the Yuzhny airport terminal complex developed by British company Twelve Architects.\\ \hline

\end{tabular*}
\bigskip
\caption{Examples of translation from CharRNN, Transformer and CharTransformer, illustrating the main error types, observed in a random sample of 100 sentences for the Russian to English language pair.}
  \label{table:Examples}
\end{table*}

\noindent  \textbf{Instability of metrics} Interestingly, we can observe a high variation in metrics (see Table \ref{table:Res}). However, it is expected due to different degree of correlation between metrics and human scores. If we rely solely on highly popular BLEU conclusions may be misleading as it is not the best metric for three out of four language pairs (see Table \ref{table:metrics}). From Table \ref{table:Res}, we can see that improvement of 1 BLEU point does not necessary lead to improvements in other metrics. Hence, we make our conclusions based on least two metrics out of three where it is possible.\\

\noindent  \textbf{RNN vs. Transformer} \citet{lee2017fully} reported a training time for CharRNN of approximately 2 weeks on a single GPU. However, we can not directly compare training time of CharRNN to our character-level models due to usage of different frameworks, GPUs, batch sizes and depth of models. From Table \ref{table:time}, we can observe that it takes roughly 38 and 25 hours to train the character-level Transformer and CharTransformer respectively. In addition, the character-level Transformer and CharTransformer show better results for all language pairs (see Table \ref{table:Res}). Hence, we train our deeper character-level models substantially faster and outperform previously obtained results by a large margin. We conclude that Transformer applied at the character level and CharTransformer are better than CharRNN.\\

\noindent  \textbf{Character-level Transformer vs. CharTransformer} According to Table  \ref{table:Res}, Transformer applied at character level is the best performer in FI-EN and RU-EN. CharTransformer shows better results in DE-EN and CS-EN. In the experiments, we do not observe superiority of CharTransformer in results over Transformer. However, CharTransformer is 34 percent faster. We conclude that CharTransformer is promising and worth further investigation.   \\

\noindent  \textbf{Character- vs. subword-level} From Table \ref{table:Res}, we can observe that character-level models in some cases outperform subword-level models. CharTransformer and character-level Transformer outperform subword-level Transformer in FI-EN. In addition, character-level Transformer shows comparable results in RU-EN and CharTransformer is slightly worse in CS-EN than subword-level Transformer. The subword-level model is convincingly the best only in DE-EN. Similarly to \citet{gupta2019character}, we observe that the character-level models are competitive to the subword-level models, but do not outperform them. It shows that these models are promising and should get more attention.  \\

\subsection{Qualitative Analysis}
We have performed a qualitative inspection of 100 randomly sampled sentences from newstest-2014 of the Russian-English language pair for the four models compared (CharRNN, subword-level and character-level Transformer, and CharTransformer). We selected this language pair because of the relatively large typological distance between both languages, as well as the challenging transliteration issues that might arise from the mapping of two alphabets. Overall, CharRNN displays a clear inferiority to the Transformer architectures. The quality of CharTransformer is indeed slightly lower than the Transformers (in accordance with the quantitative results), but not much. Noteworthy are the following, persisting error categories (referencing examples a--d drawn from Table~\ref{table:Examples}): \\

\noindent  \textbf{Entities and transliteration} Named entities, especially proper nouns, are a classic hindrance in MT, especially when source and target language use a different alphabet. All systems suffer from artifacts in this area, but CharRNN most heavily. In many cases, systems propose entirely different transliterations of the proper nouns in the source language (a). \\

\noindent  \textbf{Length-related artifacts} CharRNN translations often feature the unnecessary repetitions of chunks (`flooding'), as well as incomplete words (b). Likewise, CharRNN often produces  incorrect syntactic constructions which is rare with the other architectures. Overall, the Transformers yield slightly more concise translations than the CharTransformer (121.58$\pm$59.92 (bpe) vs. 125.18$\pm$64.80 (char) vs. 126.11$\pm$64.94 characters on average) (d), which might be related to the settings of the beam search. \\

\noindent  \textbf{Fixed expressions} In comparison to the Transformer architectures, CharRNN sometimes struggle to translate figurative language use and idiomatic expressions. The same is true for the CharTransformer, but to a lesser extent (c). \\

\noindent \textbf{Overall quality} We conclude that CharRNN is relatively less capable of modelling longer-range sequences at the character level. To the human eye, and however small the sample size, the differences between the Transformers and CharTransformer are limited, although the Transformers generally yields minimalist translations, that are of a slightly higher quality.

\section{Conclusion and Future Work}
In this work, we applied Transformer from OpenNMT-py at character level and proposed a new character-level Transformer-based NMT architecture, CharTransformer. We evaluated it on four languages from WMT’15 corpora and compared these models to the character-level architecture previously proposed by \citet{lee2017fully}. We showed that character-level Transformer and CharTransformer outperform this model in all tasks. We demonstrated that character-level translation does not require weeks of training and expensive multi GPU training scheme anymore to strong results. In addition, we showed that CharTransformer performs comparably with character-level Transformer and is 34 percent faster. CharTransformer outperforms the subword-level model in FI-EN and shows competitive results in CS-EN. We conclude that both models are promising for character-level translation and can stimulate further research in this field. 
 
We provide the repository\footnote{The link will be provided later} that contains the source code of the implemented models. In future research, we would like to investigate multilingual character-level translation with Transformer and CharTransformer. In addition, we will research different properties of these models. Finally, we should emphasize that our results that we might close the gap between character-level and subword-level NMT in a very near future.

\bibliography{anthology,acl2020}
\bibliographystyle{acl_natbib}

\end{document}